%% file: main.tex
\documentclass[
]{ceurart}

\usepackage[utf8]{inputenc}
\usepackage{graphicx}
\usepackage{amsmath, amssymb}
\usepackage{hyperref}
\usepackage{listings}
\usepackage[numbers]{natbib}
\usepackage{placeins} 

\begin{document}

\copyrightyear{2025}
\copyrightclause{Copyright for this paper by its authors.
Use permitted under Creative Commons License Attribution 4.0
International (CC BY 4.0).}


\conference{{D}e{F}actify 4.0: Fourth workshop on Multimodal Fact-Checking and Hate Speech Detection, March 2025, Philadelphia, Pennsylvania, USA}

\title{Findings of the Counter Turing Test: AI-Generated Image Detection}

\tnotemark[1]
\tnotetext[1]{This document is based on the CEUR-WS template and incorporates topics inspired by the Defactify workshop series.}

\author[1]{Rajarshi Roy}[email=royrajarshi0123@gmail.com]
\author[2]{Nasrin Imanpour}
\author[3]{Ashhar Aziz}
\author[4]{Shashwat Bajpai}
\author[5]{Gurpreet Singh}
\author[6]{Shwetangshu Biswas}
\author[7]{Kapil Wanaskar}
\author[8]{Parth Patwa}
\author[9]{Subhankar Ghosh}
\author[10]{Shreyas Dixit}
\author[1]{Nilesh Ranjan Pal}
\author[2]{Vipula Rawte}
\author[2]{Ritvik Garimella}
\author[13]{Amitava Das}
\author[2]{Amit Sheth}
\author[11]{Vasu Sharma}
\author[12]{Aishwarya Naresh Reganti}
\author[11]{Vinija Jain}
\author[12]{Aman Chadha}

\address{$^1$Kalyani Government Engineering College, India. $^2$University of South Carolina, USA. $^3$IIIT Delhi, India. $^4$BITS Pilani Hyderabad Campus, India. $^5$IIIT Guwahati, India. $^6$NIT Silchar, India. $^7$San José State University, USA. $^8$UCLA, USA. $^9$Washington State University, USA. $^{10}$Vishwakarma Institute of Information Technology, India. $^{11}$Meta AI, USA. $^{12}$Amazon AI, USA. $^{13}$BITS Pilani Goa, India.}

\begin{abstract}
The rapid advancements in generative AI technologies, such as Stable Diffusion, DALL-E, and Midjourney, have significantly transformed the creation of synthetic visual content. While these models enable innovation across industries, they also pose serious challenges, including misinformation, disinformation, and biased content generation. The increasing realism of AI-generated images makes their detection a pressing concern for researchers, policymakers, and industry stakeholders.

In this paper, we present the findings of the Defactify 4.0 workshop, which introduced the Counter Turing Test (CT2) for AI-Generated Image Detection. The competition consisted of two key tasks: (1) binary classification of images as either AI-generated or real and (2) identification of the specific generative model responsible for an AI-generated image. To support both tasks, we employed the MS COCOAI dataset, a benchmark of 96000 real and synthetic images generated by five state-of-the-art models alongside real images from MS COCO.

Participants employed diverse detection strategies, including convolutional neural networks (CNNs), Vision Transformers (ViTs), frequency-based analysis, contrastive learning, and multimodal techniques. The results demonstrated that while AI-generated images can be detected with high accuracy (F1-score > 0.83), identifying the exact model used remains significantly more challenging (highest F1-score: 0.4986). These findings highlight the need for improved model fingerprinting, adversarial robustness, and real-time detection mechanisms.

This study contributes to the development of scalable and reliable detection systems, bridging the gap between academic research and practical applications in safeguarding digital ecosystems against AI-driven misinformation. Future work should focus on enhancing adversarial resilience, refining model attribution techniques, and leveraging multimodal approaches to strengthen AI-generated content detection.\end{abstract}


\maketitle

\input{introduction}

\input{related_work}

\input{shared_task}
\input{participating_systems}
\input{results}
\input{conclusion}

\bibliography{references} 
\end{document}

%% file: introduction.tex
\section{Introduction}

Generative AI tools such as Stable Diffusion \cite{rombach2022high}, DALL-E \cite{ramesh2021zero}, and  Midjourney \cite{midjourney_website} have produced a step-change in the accessibility of synthetic image creation. Unlike earlier GAN-based systems, modern diffusion models and text-to-image architectures can produce photorealistic outputs from simple text prompts. This enables legitimate use cases in design, entertainment, and scientific visualization, while simultaneously lowering the barrier for malicious misuse.

The consequences of this misuse are tangible. In one widely reported incident, an AI-generated image falsely depicting an explosion near the Pentagon circulated on social media and briefly triggered a measurable decline in U.S. equity markets \cite{apnews_pentagon_ai_2023}. In another case documented during the 2020 U.S. presidential election cycle, manipulated and AI-generated media were increasingly recognized as tools for spreading political disinformation, including fabricated videos and imagery designed to mislead voters, inflame polarization, and erode public trust in democratic processes \cite{cnbc_deepfakes_election_2019}.


A core technical challenge underlies all of these cases: state-of-the-art generative models including Stable Diffusion XL \cite{podell2023sdxlimprovinglatentdiffusion}, DALL-E 3 \cite{BetkerImprovingIG}, and Midjourney 6 \cite{midjourney_website} have pushed synthetic image quality to the point where human observers and many classical detection methods are routinely deceived. This creates urgency at multiple levels: for researchers developing detection algorithms, for policymakers drafting regulatory frameworks, and for platforms responsible for moderating content at scale.


This paper presents the findings of the Counter Turing Test (CT2) for AI-Generated Image Detection, hosted as part of the Defactify 4.0 workshop \footnote{\url{https://defactify.com/}} . The shared task was designed to benchmark the state of the art on two distinct problems: binary discrimination of real versus synthetic images, and fine-grained attribution of synthetic images to the specific model that produced them. Our analysis of participating systems, their methodologies, and their results is intended to characterize where the field currently stands and where the hardest unsolved problems lie. This work builds directly on a companion contribution from our group, A Comprehensive Dataset for Human vs. AI Generated Image Detection \cite{roy2026comprehensivedatasethumanvs}, which describes the benchmark used throughout the competition.



%% file: related_work.tex
\section{Related Work}

With the rapid advancement of generative models, identifying synthetic content has become a critical area of research. Recent literature focuses on leveraging the unique structural, semantic, and statistical artifacts left behind by Generative Adversarial Networks (GANs) \cite{goodfellow2014generativeadversarialnetworks} and Diffusion Models (DMs).

\textbf{Reconstruction and Error-Based Methods: \\} 
A prominent line of research exploits the premise that generative models can reconstruct synthetic images with higher fidelity than real-world photographs. DIRE \cite{Wang_2023_ICCV} introduces a representation based on reconstruction error, noting that diffusion models more accurately reconstruct images within their own training distribution. Similarly, AEROBLADE \cite{ricker2024aerobladetrainingfreedetectionlatent} approaches detection without any training overhead, instead exploiting the reconstruction behavior of the autoencoder component in latent diffusion models to identify synthetic images at inference time. To improve efficiency and accuracy, LaRE2 \cite{Luo_2024_CVPR} shifts this analysis to the latent space, while DRCT \cite{pmlr-v235-chen24ay} employs contrastive training on reconstruction pairs to learn generator-invariant features that improve cross-model generalization.

\textbf{Foundation Models and Vision-Language Frameworks: \\}
Another approach leverages the robust, semantic feature spaces of large-scale pre-trained models. UnivFD \cite{ojha2024universalfakeimagedetectors} demonstrates that frozen CLIP-ViT features can serve as a universal detector across diverse model families. RIGID \cite{he2024rigidtrainingfreemodelagnosticframework} identifies AI-generated images by exploiting the elevated sensitivity of DINOv2 image representations to structured noise perturbations, it is a property that distinguishes synthetic images from real ones in the feature space. Furthermore, research has integrated textual cues to guide visual forensics; C2P-CLIP \cite{tan2024c2pclipinjectingcategorycommon} injects category-related concepts via category common prompts to improve detection capabilities, while LASTED \cite{wu2025generalizablesyntheticimagedetection} optimizes forensic feature extraction through language-guided contrastive learning.

\textbf{Statistical and Noise Fingerprinting: \\}
Researchers have also focused on low-level signal irregularities, such as high-frequency artifacts and sensor noise patterns. DNF \cite{zhang2025diffusionnoisefeatureaccurate} extracts "fingerprints" from the noise sequence generated during the inverse diffusion process. To simplify the detection pipeline, SSP \cite{chen2024singlesimplepatchneed} argues that generative models fail to mimic natural camera noise in "simple" image patches, making them ideal for forensic analysis. Additionally, LGrad \cite{Tan_2023_CVPR} reframes detection as a model-dependent rather than data-dependent problem by converting images into their gradient representations, which expose generator-specific artifacts more reliably than pixel-level features.

\textbf{Robustness and Multi-Cue Aggregation: \\}
To combat post-processing degradations like JPEG compression or resizing, several works propose hybrid architectures. GenDet \cite{zhu2023gendetgoodgeneralizationsaigenerated} introduces a training protocol focused on universal generative artifacts to ensure state-of-the-art generalization across various benchmarks, specifically targeting features orthogonal to common image biases.





%% file: shared_task.tex
\section{Task Details}

The Defactify 4.0 workshop introduced two sub-tasks under the Counter Turing Test (CT2) for AI-generated image detection, designed to address the growing challenges posed by modern generative image models.

\subsection{Data}

The shared task was built around the MS COCOAI \cite{roy2026comprehensivedatasethumanvs} benchmark introduced in our companion dataset paper \cite{roy2026comprehensivedatasethumanvs}, to which we refer readers for full construction details. Briefly, the dataset pairs real images from MS COCO \cite{lin2014microsoft} with 96000 synthetic and real images generated by prompting five contemporary text-to-image models, including Stable Diffusion 3  (SD3 ) \cite{esser2024scalingrectifiedflowtransformers}, Stable Diffusion XL (SDXL) \cite{podell2023sdxlimprovinglatentdiffusion}, Stable Diffusion 2.1 (SD 2.1) \cite{rombach2022high}, DALL-E 3 \cite{BetkerImprovingIG}, and Midjourney 6 \cite{midjourney_website}, by using the original MS COCO captions as input prompts. This design ensures that real and synthetic images are semantically comparable, isolating visual generation artifacts rather than content differences.


Each sample is enriched with annotated metadata, including the generative model used, input captions, prompts, and observable artifacts such as spectral inconsistencies or edge distortions. 

Data was partitioned into train, validation, and test splits of 42,000 / 9,000 / 45,000 images, respectively. The test split is substantially larger than typical benchmarks, reflecting deliberate design choices described in \cite{roy2026comprehensivedatasethumanvs} to stress-test generalization.


\subsection{Tasks}

\begin{itemize}

    \item \textbf{Task A: Binary Classification} \newline
    Participants were required to determine whether a given image was AI-generated or captured in the real world.

    \item \textbf{Task B: Model Identification} \newline
    Building on Task A, this task required participants to identify the specific generative model responsible for creating a given image. The target models included Stable Diffusion 3 (SD 3) \cite{esser2024scalingrectifiedflowtransformers}, Stable Diffusion XL (SDXL) \cite{podell2023sdxlimprovinglatentdiffusion}, Stable Diffusion 2.1 (SD 2.1) \cite{rombach2022high}, DALL-E 3 \cite{BetkerImprovingIG}, and Midjourney 6 \cite{midjourney_website}.

\end{itemize}

Additionally, artificial fingerprint analysis was conducted to investigate model-specific traces, including spectral inconsistencies and anomalous auto-correlation patterns, which may help distinguish between different generative models.

\subsection{Evaluation}

Performance in the competition is assessed using the \textbf{F$_1$-score}. 
For \textbf{Task A}, we report the \textbf{weighted F$_1$-score}, which accounts for label imbalance by averaging the F$_1$-scores of each class weighted by their support. 
For \textbf{Task B}, we use the \textbf{macro F$_1$-score}, which treats all classes equally by computing the unweighted mean of the per-class F$_1$-scores, thus emphasizing the ability to distinguish unique patterns across different model-generated outputs.







\subsection{Baseline}  

To establish a baseline, we trained a ResNet-50 classifier using image representations in the frequency domain. These frequency domain representations were generated by applying a preprocessing strategy inspired by the methodology presented in \cite{Corvi_2023_CVPR}. This transformation captures global frequency characteristics that help reveal subtle artifacts often present in synthetic images.

The overall pipeline is illustrated in Figure~\ref{fig:baseline_pipeline}. Starting with an input image, we convert it into its frequency domain using a 2D Fourier Transform. The resulting representation is then fed into a ResNet-50 CNN \cite{he2016deep} model trained to classify the image as either "Real" or "Fake."

This baseline allows us to assess the effectiveness of frequency-based features in distinguishing between real and synthetic imagery, serving as a point of comparison for more sophisticated techniques.

\begin{center}
    \includegraphics[width=0.9\textwidth]{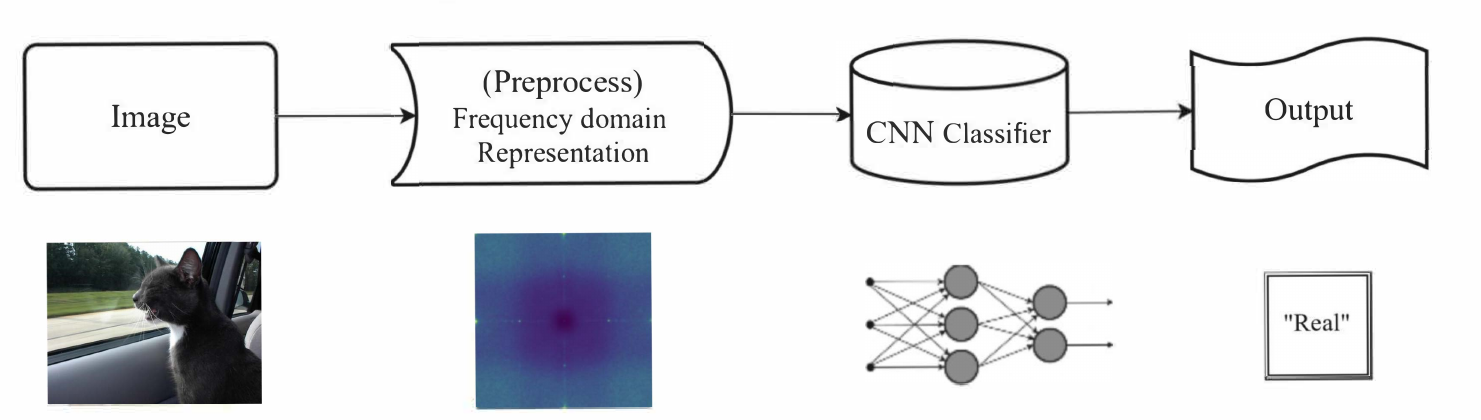} 
    \captionof{figure}{\textbf{Baseline workflow.} The input image is first transformed into its frequency domain representation and then passed through a ResNet-50 CNN  \cite{he2016deep} classifier to predict whether it is real or fake.}
    \label{fig:baseline_pipeline}
\end{center}  

%% file: participating_systems.tex
\section{Participating Systems}  

The AI-generated image detection shared task attracted significant interest, with over 57 registrations on the competition web page. Ultimately, 10 teams submitted results to the final leaderboard, and 7 teams provided paper submissions.

The first of which is Team \textbf{NYCU} \cite{yang2025team} ranked in the top three in Task A, employing a combination of CNN and CLIP-ViT classifiers. Their CNN classifier used EfficientNet-B0 as the backbone, incorporating RGB channels, frequency features, and reconstruction errors. Additionally, they utilized a CLIP-ViT-based approach, where a pretrained CLIP image encoder extracted image features, followed by an SVM classifier for detection.  

Team \textbf{SKDU} \cite{malviya2025skdudefactify40vision} enhanced AI-generated image detection using a fine-tuned Vision Transformer (ViT) with advanced data augmentation. Their approach applies additional perturbations (flipping, rotation, noise, compression) to boost robustness. 

Team \textbf{TAHAKOM} \cite{albusayes_beyond_rgb} introduced RSID (Robust Synthetic Image Detection), a framework utilizing alternative image color spaces (Lab and YCbCr) and multi-modal contrastive learning. To further improve robustness, they experimented with multiple inference strategies, including image–text similarity, nearest neighbors, and random forests, and employed extensive data augmentation. Their approach achieved an 83.05\% F1-score in the Task A.

Team \textbf{RoVIT} \cite{penades_analyzing_cnns_defactify} addressed the challenge by leveraging spatial and frequency characteristics of the images. Their approach involved image preprocessing with resizing and statistical normalization. By experimenting with different image sizes and model architectures, they found that larger image sizes and preservation of spatial
domain integrity led to superior performance in AI-generated image detection.  

Team \textbf{Xiaoyu} \cite{guo2026nau} developed a multi-modal multi-task model integrating pre-trained BERT and CLIP Vision encoders. Their approach leverages cross-modal feature fusion together with a multi-task learning objective, and incorporates pseudo-labeling–based data augmentation to expand the training dataset. This design optimizes performance for both binary and multi-class classification tasks, achieving fifth place in both Task A and Task B.

Team \textbf{Dakiet} \cite{duong2025scalableframeworkclassifyingaigenerated} ranked second in Task A and third in Task B, developing a scalable framework integrating perceptual hashing, similarity measurement, and pseudo-labeling. Their method leveraged the Swin Transformer V2 Base as the backbone model and incorporated contrastive learning for incremental adaptation. Notably, their approach allowed adaptation to new generative models without requiring retraining, making it highly flexible.  

Team \textbf{Nitiz} \cite{khanal_attentionfusion} introduced AttentionFusion, a multi-stream neural architecture for AI image detection that combined texture, color, and shape analysis. The model incorporated EfficientNet, ResNet34, and Vision Transformer, with an attention mechanism to dynamically weigh feature contributions. This approach enabled effective binary and multi-class classification, improving AI-generated image detection and source identification.  
Notably, the approach emphasizes \textit{parameter-efficient feature extraction and fusion}, allowing the model to capture fine-grained visual patterns of different AI image generation models

These diverse approaches demonstrate the innovative methods teams employed to tackle the challenge of AI-generated image detection, ranging from transformer-based architectures and multi-modal learning to frequency-based analysis and advanced data augmentation strategies.

%% file: results.tex
\section{Results}

\begin{table}[h]
    \centering
    \begin{tabular}{cccc}
        \toprule
        \textbf{S.No} & \textbf{Name}         & \textbf{Team Name} & \textbf{Scores} \\
        \midrule
        1  & \textbf{Suril}                & \textbf{SeeTrails}   & \textbf{0.8334}   \\
        2  & Duong Anh Kiet       & Dakiet      & 0.8330   \\
        3  & Tsan-Tsung            & NYCU        & 0.8329   \\
        4  & Shaurya              & random.py   & 0.8326   \\
        5  & Xiaoyu               & Xiaoyu      & 0.8316   \\
        6  & Raghad\_Khalid       & TAHAKOM     & 0.8305   \\
        7  & Shrikant Malviya     & SKDU        & 0.8292   \\
        8  & Nitiz Khanal         & Nitiz       & 0.8152   \\
           -        &-              & \textbf{BASELINE}    & \textbf{0.80144}  \\
        9  & Omar Nasr            & OAR         & 0.7996   \\
        10 & Hector               & RoVIT       & 0.7590   \\
        \bottomrule
    \end{tabular}    \caption{Leaderboard for Task A: Classify each image as either AI-generated or created by a human.}
    \label{tab:task_a}
\end{table}

Table \ref{tab:task_a} showcases the leaderboard for Task A, where participants are ranked based on their scores. The highest score of 0.8334 is achieved by Suril from Team SeeTrails, followed closely by Duong Anh Kiet from Team Dakiet with a score of 0.8330. The top five participants have scores above 0.83, demonstrating strong performance in this task.


\begin{table}[h]
    \centering
    \begin{tabular}{cccc}
        \toprule
        \textbf{S.No} & \textbf{Name}         & \textbf{Team Name} & \textbf{Scores} \\
        \midrule
        1  & \textbf{Suril}                & \textbf{SeeTrails}   & \textbf{0.4986}   \\
        2  & Shaurya              & random.py   & 0.4936   \\
        3  & Duong Anh Kiet       & Dakiet      & 0.4935   \\
        4  & Tsan-Tsung            & NYCU        & 0.4910   \\
        5  & Xiaoyu               & Xiaoyu      & 0.4888   \\
        6  & Shrikant Malviya     & SKDU        & 0.4864   \\
        7  & Raghad\_Khalid       & TAHAKOM     & 0.4816   \\
        -   &-                      & \textbf{BASELINE}    & \textbf{0.44913}  \\
        8  & Hector               & RoVIT       & 0.4222   \\
        9  & Nitiz Khanal         & Nitiz       & 0.4193   \\
        10 & Omar Nasr            & OAR         & 0.2726   \\
        \bottomrule
    \end{tabular}
    \caption{Leaderboard for Task B: Given an AI-generated Image, determine which specific model produced it (e.g., SD 3, SDXL, SD 2.1, DALL-E 3, or Midjourney 6).}
    \label{tab:task_b}
\end{table}

\FloatBarrier

Table \ref{tab:task_b} presents the leaderboard for Task B, which has a different ranking compared to Task A. The highest score in Task B is 0.4986, achieved by Suril from Team SeeTrails, followed closely by Shaurya from Team random.py with a score of 0.4936. The scores in Task B are generally lower, suggesting that it is a more challenging task.

%% file: conclusion.tex
\section{Conclusion}

This paper has reported the results of the Counter Turing Test (CT2) for AI-Generated Image Detection, the shared task component of the Defactify 4.0 workshop. Across ten competing teams, we observed a consistent and interpretable pattern: binary classification of AI-generated images is now a tractable problem, with the top system reaching an F1-score of 0.8334 and most competitive submissions exceeding 0.83. In contrast, determining which generative system produced a given image remains largely unsolved, with the best result at 0.4986.

This gap between the two tasks is not merely a matter of difficulty; it reflects a structural property of current generative models. The artifacts that make an image detectable as synthetic are increasingly shared across model families, while the fine-grained signatures that distinguish one model from another are subtle, inconsistent, and vulnerable to post-processing. Closing this gap will require advances in model fingerprinting, training strategies that explicitly target inter-model discrimination, and datasets with denser adversarial coverage than currently exist.

Future research should focus on enhancing adversarial robustness, scaling real-time detection systems, and integrating multimodal approaches to mitigate the risks associated with AI-generated misinformation.
